\documentclass[11pt,a4paper]{article}
\usepackage[hyperref]{acl2018}
\usepackage{times}
\usepackage{latexsym}

\usepackage{url}

\usepackage{graphicx}  

\usepackage{caption}
\usepackage{chngpage}
\usepackage{subfigure}
\usepackage{array}
\usepackage{multirow}
\usepackage{amsmath}
\usepackage{CJK}
\usepackage{amsfonts}
\usepackage{bm}
\usepackage{enumerate}
\newcommand{\ignore}[1]{}
\aclfinalcopy 

\title{Attention Focusing for Neural Machine Translation \\ by Bridging Source and Target Embeddings}

\author{Shaohui Kuang$^1$\hspace{0.5cm} Junhui Li$^1$\hspace{0.5cm}  Ant{\'o}nio Branco$^2$\hspace{0.5cm} Weihua Luo$^3$\hspace{0.5cm}  Deyi Xiong$^1$\thanks{ \hspace{0.1cm} Corresponding author}\\ 
$^1$School of Computer Science and Technology, Soochow University, Suzhou, China\\
{\tt shaohuikuang@foxmail.com, \{lijunhui, dyxiong\}@suda.edu.cn} \\
$^2$University of Lisbon, NLX-Natural Language and Speech Group,  Department of Informatics\\ 
 Faculdade de Ci{\^e}ncias,  Campo Grande, 1749-016 Lisboa, Portuga \\
 {\tt antonio.branco@di.fc.ul.pt} \\
 $^3$Alibaba Group, Hangzhou, China \\
{\tt weihua.luowh@alibaba-inc.com}
}

\date{}

\begin{document}
\begin{CJK}{UTF8}{gbsn}
\maketitle
\begin{abstract}
In neural machine translation, a source sequence of words is encoded into a vector from which a target sequence is generated in the decoding phase. Differently from statistical machine translation, the associations between source words and their possible target counterparts are not explicitly stored. Source and target words are at the two ends of a long information processing procedure, mediated by hidden states at both the source encoding and the target decoding phases. This makes it possible that a source word is incorrectly translated into a target word that is not any of its admissible equivalent counterparts in the target language. 

In this paper, we seek to somewhat shorten the distance between source and target words in that procedure, and thus strengthen their association, by means of a method we term bridging source and target word embeddings. We experiment with three strategies: (1) a source-side bridging model, where source word embeddings are moved one step closer to the output target sequence; (2) a target-side bridging model, which explores the more relevant source word embeddings for the prediction of the target sequence; and (3) a direct bridging model, which directly connects source and target word embeddings seeking to minimize errors in the translation of ones by the others. 

Experiments and analysis presented in this paper demonstrate that the proposed bridging models are able to significantly improve quality of both sentence translation, in general, and alignment and translation of individual source words with target words, in particular.
\end{abstract}

\begin{figure}[!t]
\centering
\includegraphics*[height=1.3in,width=2.1in]{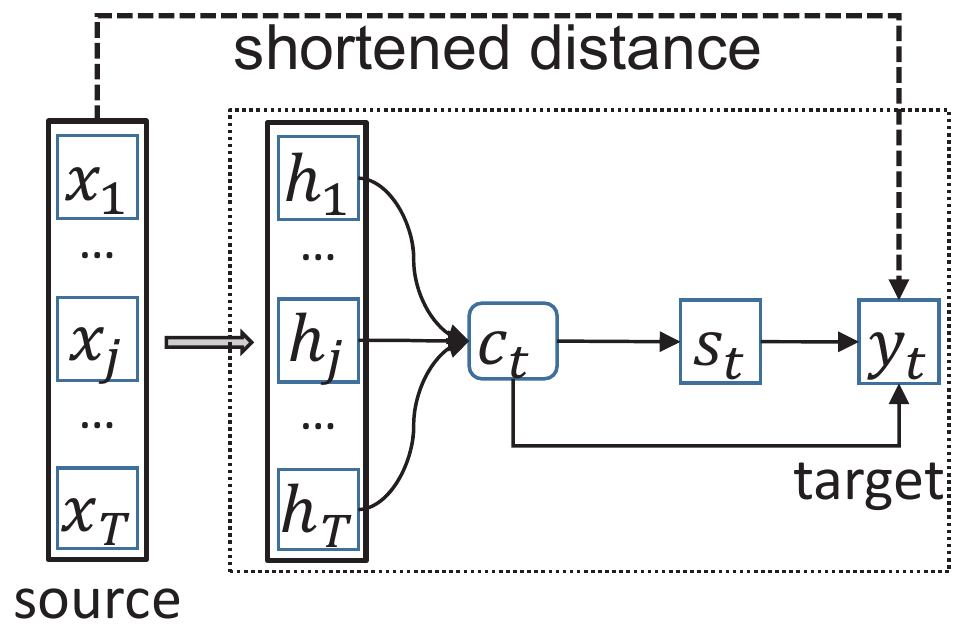}
\caption{Schematic representation of seq2seq NMT, where $x_1,\dots,x_T$ and $h_1,\dots,h_T$ represent source-side word embeddings and hidden states respectively, \(c_t\) represents a source-side context vector, \(s_t\) a target-side decoder RNN hidden state, and \(y_t\) a predicted word.
Seeking to shorten the distance between source and target word embeddings, in what we term bridging, is the key insight for the advances presented in this paper.}
\label{fig:6}
\end{figure}

\section{Introduction}
Neural machine translation (NMT) is an end-to-end approach to machine translation that has achieved  competitive results vis-a-vis statistical machine translation (SMT) on various language pairs \cite{bahdanau2015neural,cho2014learning,sutskever2014sequence,luong2015stanford}. 
In NMT, the sequence-to-sequence (seq2seq) model learns word embeddings for both source and target words synchronously. However, as illustrated in Figure~\ref{fig:6}, source and target word embeddings are at the two ends of a long information processing procedure.  The individual associations between them will gradually become loose due to the separation of source-side hidden states (represented by $h_1,\dots,h_T$ in Fig.~\ref{fig:6}) and a target-side hidden state (represented by $s_t$ in Fig.~\ref{fig:6}). As a result, in the absence of a more tight interaction between source and target word pairs, the seq2seq model in NMT produces tentative translations that contain incorrect alignments of source words with target counterparts that are non-admissible equivalents in any possible translation context. 

Differently from SMT, in NMT an attention model is adopted to help align output with input words. The attention model is based on the estimation of a probability distribution over all input words for each target word. Word alignments with attention weights can then be easily deduced from such distributions and support the translation. 
Nevertheless, sometimes one finds translations by NMT that contain surprisingly wrong word alignments, that would unlikely occur in SMT. 

\begin{figure}[!t]
\centering
\includegraphics*[height=1.1in,width=2.8in]{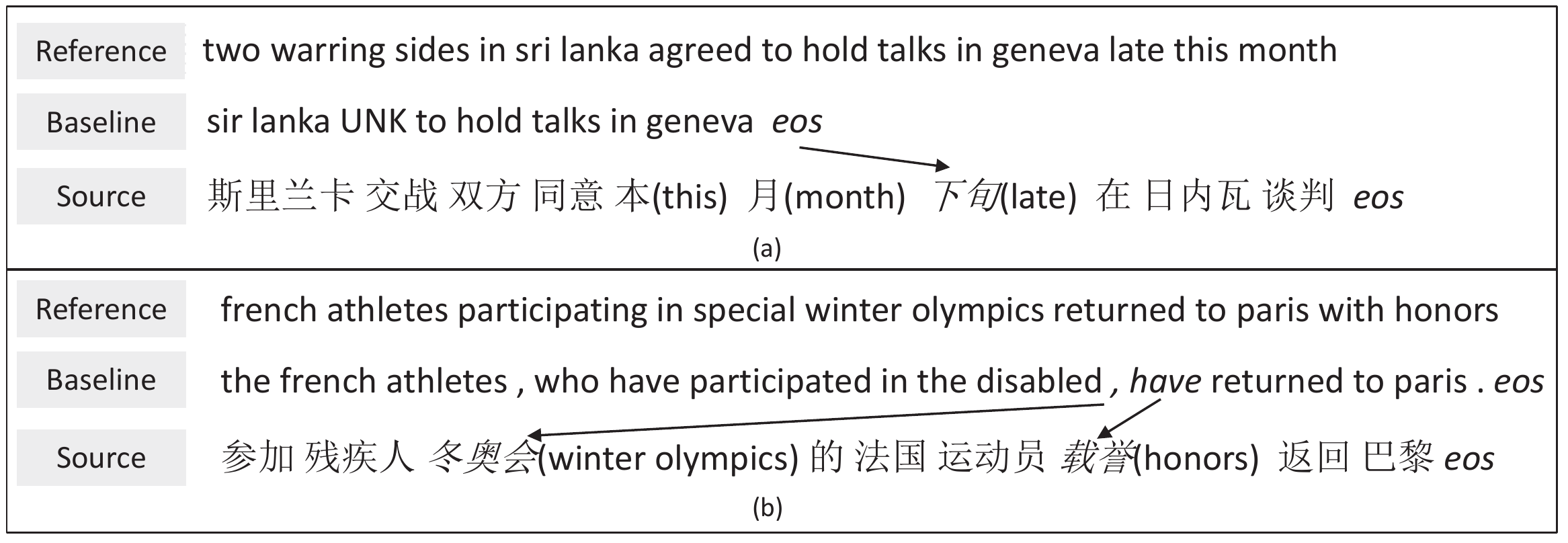}
\caption{Examples of NMT output with incorrect alignments of source and target words that cannot be the translation of each other in any possible context.}
\label{fig:8}
\end{figure}

For instance, Figure 2 shows two Chinese-to-English translation examples by NMT. In the top example, the NMT seq2seq model incorrectly aligns the target side end of sentence mark \textit{eos} to \textit{下旬/late} with a high attention weight (0.80 in this example) due to the failure of appropriately capturing the similarity, or the lack of it, between the source word \textit{下旬/late} and the target \textit{eos}.
It is also worth noting that, as \textit{本/this} and \textit{月/month} end up not being translated in this example, inappropriate alignment of target side \textit{eos} is likely the responsible factor for under translation in NMT as the decoding process ends once a target \textit{eos} is generated. 
Statistics on our development data show that as much as 50\% of target side \textit{eos} do not properly align to source side \textit{eos}.

The second example in Figure 2 shows another case where source words are translated into target items that are not their possible translations in that or in any other context. In particular, \textit{冬奥会/winter olympics} is incorrectly translated into a target comma ``,'' and \textit{载誉/honors} into \textit{have}. 


In this paper, to address the problem illustrated above, we seek to shorten the distance within the  seq2seq NMT information processing procedure between source and target word embeddings. This is a method we term as bridging, and can be conceived as strengthening the focus of the attention mechanism into more translation-plausible source and target word alignments.
In doing so, we hope that the seq2seq model is able to learn more appropriate word alignments between source and target words. 

We propose three simple yet effective strategies to bridge between word embeddings. The inspiring insight in all these three models is to move source word embeddings closer to target word embeddings along the seq2seq NMT information processing procedure. We categorize these strategies in terms of how close the source and target word embeddings are along that procedure, schematically depicted in Fig. 1.  

\begin{enumerate}[(1)]
\item \textbf{Source-side bridging model:}  Our first strategy for bridging, which we call source-side bridging, is to move source word embeddings just one step closer to the target end. Each source word embedding is concatenated with the respective source hidden state at the same position so that the attention model can more closely benefit from source word embeddings to produce word alignments.

\item \textbf{Target-side bridging model:} In a second more bold strategy, we seek to incorporate relevant source word embeddings more closely into the prediction of the next target hidden state. In particular, the most appropriate source words are selected according to their attention weights and they are made to more closely interact with target hidden states.

\item \textbf{Direct bridging model:} The third model consists of directly bridging between source and target word embeddings. The training objective is optimized towards minimizing the distance between target word embeddings and their most relevant source word embeddings, selected  according to the attention model.
\end{enumerate}

Experiments on Chinese-English translation with extensive analysis demonstrate that directly bridging word embeddings at the two ends can produce better word alignments and thus achieve better translation.

\section{Bridging Models}

As suggested by Figure 1, there may exist different ways to bridge between $x$ and $y_t$. We concentrate on the folowing three bridging models.


\begin{figure}[!t]
\centering
\includegraphics*[height=1.1in,width=1.0in]{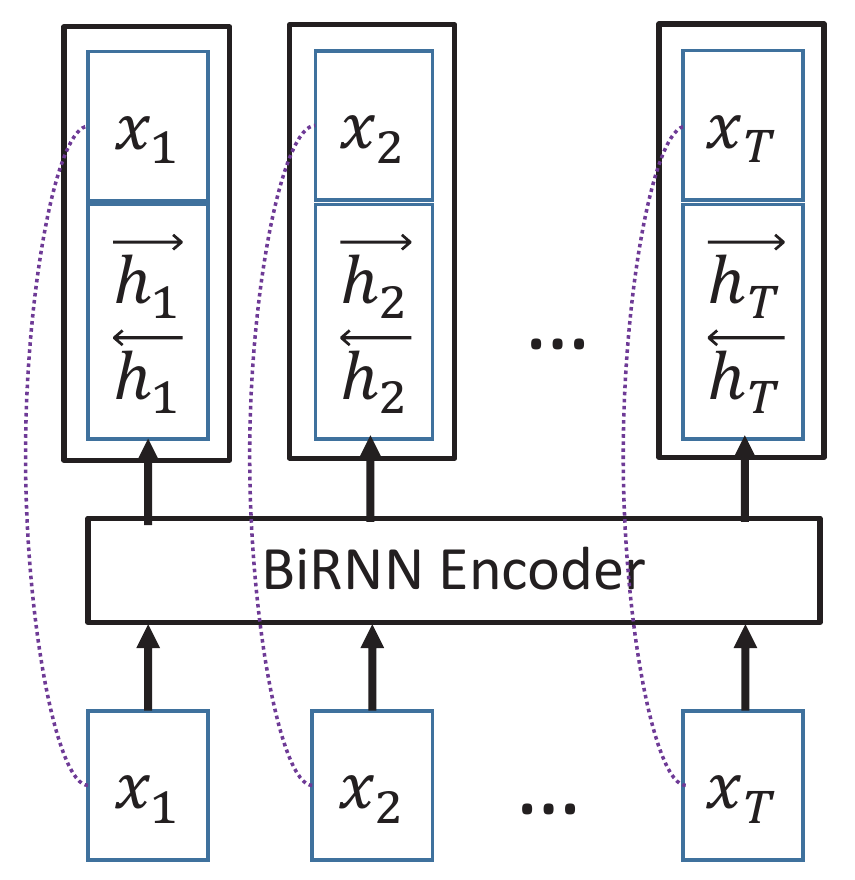}
\caption{Architecture of the source-side bridging model.}
\label{fig:2}
\end{figure}

\subsection{Source-side Bridging Model}
Figure~\ref{fig:2} illustrates the source-side bridging model. The encoder reads a word sequence equipped with word embeddings and generates a word annotation vector for each position. Then we simply concatenate the word annotation vector with its corresponding word embedding as the final annotation vector. For example, the final annotation vector \(h_j\) for the word \(x_j\) in Figure~\ref{fig:2} is \([\overrightarrow{h_j};\overleftarrow{h_j};x_j]\), where the first two sub-items \([\overrightarrow{h_j};\overleftarrow{h_j}]\) are the source-side forward and backward hidden states and \(x_j\) is the corresponding word embedding. In this way, the word embeddings will not only have a more strong contribution in the computation of attention weights, but also be part of the annotation vector to form the weighted source context vector and consequently have a more strong impact in the prediction of target words.

\begin{figure}[!t]
\centering
\includegraphics*[height=1.2in,width=1.8in]{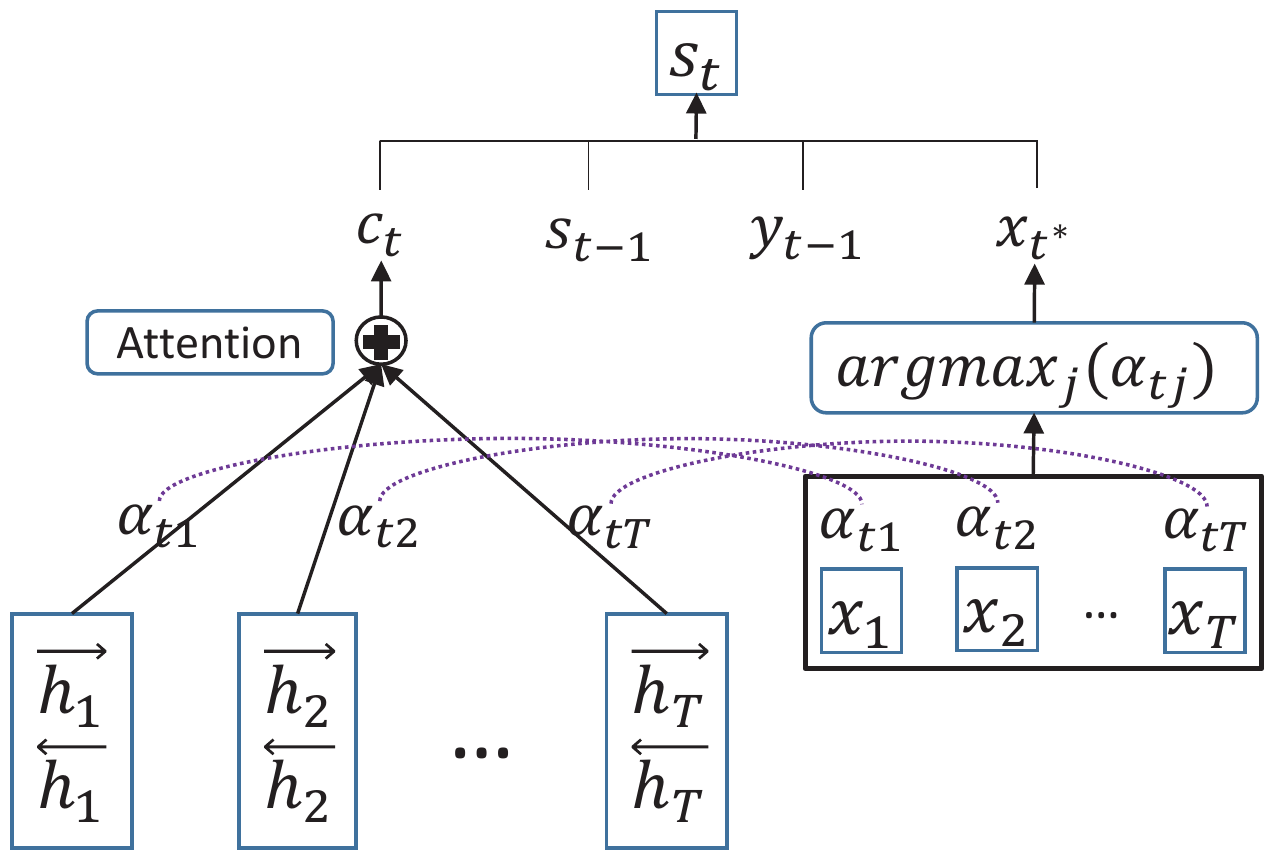}
\caption{Architecture of target-side bridging model.}
\label{fig:3}
\end{figure}

\subsection{Target-side Bridging Model}

While the above source-side bridging method uses the embeddings of all words for every target word, in the target-side method only more relevant source word embeddings for bridging are explored, rather than all of them. This is partially inspired by the word alignments from SMT, where words from the two ends are paired as they are possible translational equivalents of each other and those pairs are explicitly recorded and enter into the system inner workings. In particular, for a given target word, we explicitly determine the most likely source word aligned to it and use the word embedding of this source word to support the prediction of the target hidden state of the next target word to be generated. 

Figure~\ref{fig:3} schematically illustrates the target-side bridging method, where the input for computing the hidden state \(s_t\) of the decoder is augmented by $x_{t^*}$, as follows:

\begin{equation}
s_t = f(s_{t-1}, y_{t-1}, c_t, x_{t^*})
\end{equation}
where $x_{t^*}$ is the word embedding of the selected source word which has the highest attention weight:

\begin{equation}
t^* = \operatorname{arg\,max}_{j}(\alpha_{tj})
\end{equation}
where \(\alpha_{tj}\) is the attention weight of each hidden state \(h_j\) computed by the attention model

\begin{figure}[!t]
\centering
\includegraphics*[height=1.3in,width=1.7in]{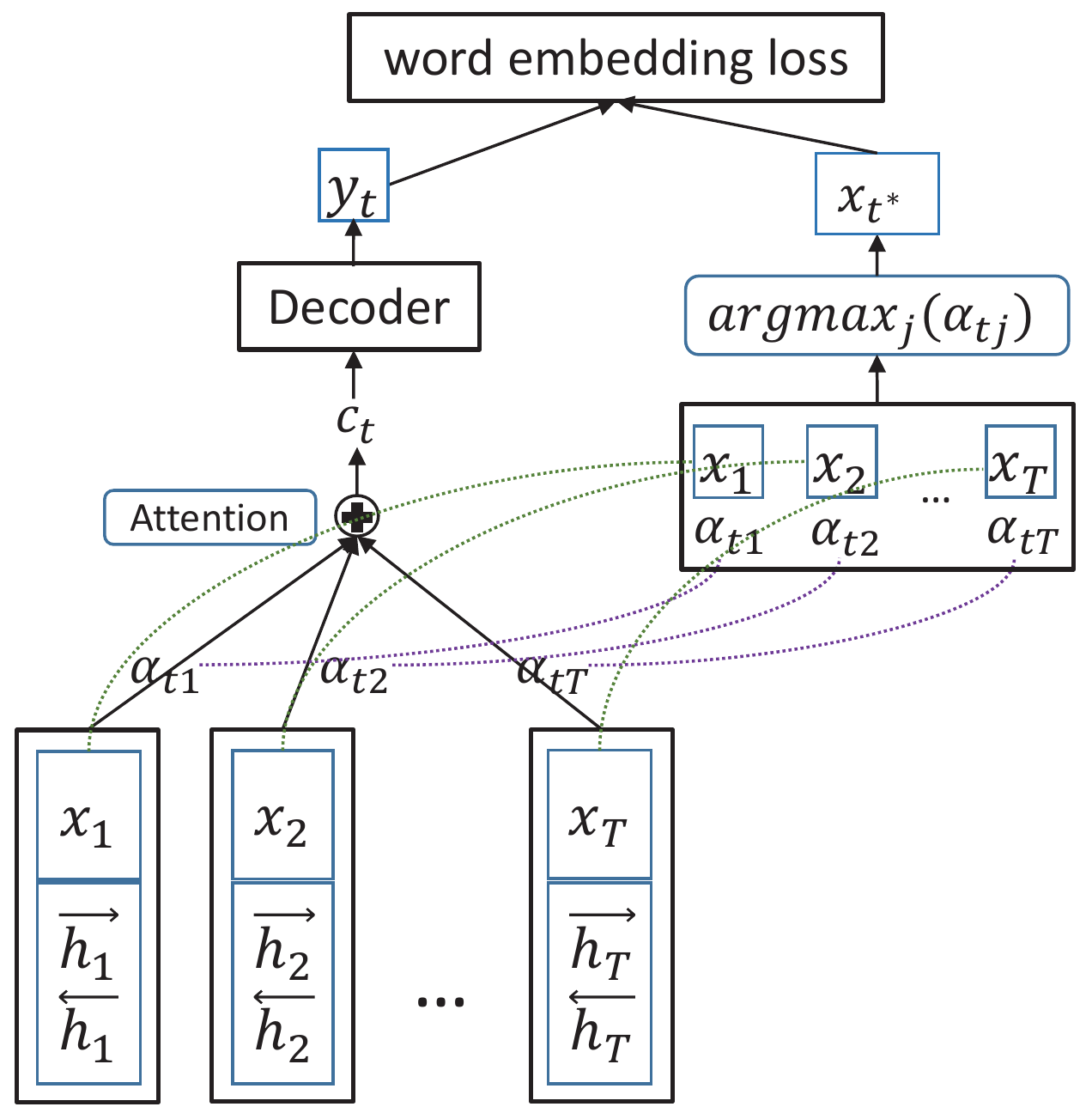}
\caption{Architecture of direct bridging model.}
\label{fig:4}
\end{figure}

\subsection{Direct Bridging Model}
Further to the above two bridging methods, which use source word embeddings to predict target words, we seek to bridge the word embeddings of the two ends in a more direct way. This is done by resorting to an auxiliary objective function to narrow the discrepancy between word embeddings of the two sides. 

Figure~\ref{fig:4} is a schematic representation of our direct bridging method, with an auxiliary objective function. More specifically, the goal is to let the learned word embeddings on the two ends be transformable, i.e. if a target word $e_i$ aligns with a source word $f_j$, a transformation matrix $W$ is learned with the hope that the discrepancy of $Wx_i$ and $y_j$ tends to be zero. Accordingly, we update the objective function of training  for a single sentence with its following extended formulation:

\begin{equation}
\begin{split}
&L(\theta) = -\sum_{t=1}^{T_y} (\log p(y_t|y_{<t},x) - \|Wx_{t^*} - y_t\|^2) \\
\end{split}
\end{equation}
\ignore{\begin{equation}
\begin{split}
&\widetilde L(\theta) = \\ 
&-\frac{1}{N} \sum_{n=1}^N \sum_{t=1}^{T_y} \left\{\log p(y_t^n|y_{<t}^n,x^n) - \|Wx_{t^*}^n - y_t^n\|^2\right\}
\end{split}
\end{equation}
}
where $\log p(y_t|y_{<t},x)$ is the original objective function of the NMT model, and 
the term \(\|Wx_{t^*} - y_t\|^2\) measures and penalizes the difference between target word $y_t$ and its aligned source word $x_{t^*}$, i.e. the one with the highest attention weight, as computed in Equation 2. Similar to \citeauthor{mi2016supervised} \shortcite{mi2016supervised}, we view the two parts of the loss in Equation 3 as equally important.

At this juncture, it is worth noting the following:
\begin{itemize}
\item Our direct bridging model is an extension of the source-side bridging model, where the source word embeddings are part of the final annotation vector of the encoder. We have also tried to place the auxiliary object function directly on the NMT baseline model. However, our empirical study showed that the combined objective consistently worsens the translation quality. We blame this on that the learned word embeddings on two sides by the baseline model are too heterogeneous to be constrained.

\item Rather than using a concrete source word embedding $x_{t^*}$ in Equation 3, we could also use a weighted sum of source word embeddings, i.e. $\sum_{j}\alpha_{tj}h_j$. However, our preliminary experiments showed that the performance gap between these two methods is very small. Therefore, we use $x_{t^*}$ to calculate the new training objective as shown in Equation 3 in all experiments.
\end{itemize}

\section{Experiments}

As we have presented above three different methods to bridge between source and target word embeddings, in the present section we report on a series of experiments on Chinese to English translation that are undertaken to assess the effectiveness of those bridging methods. 

\begin{table*}[!t]
\centering
\small
\begin{tabular}{l|l|lllll|ll}
\hline 
& \bf Model & \bf NIST06 & \bf NIST02 & \bf NIST03 & \bf NIST04  & \bf NIST08 & \bf Avg\\ 
\hline
\multirow{5}{*}{BLEU} & cdec (SMT)               & 34.00 & 35.81 & 34.70  & 37.15 & 25.28 & 33.23 \\
\cline{2-8}
&RNNSearch*         & 35.92 & 37.88  & 36.21 & 38.83  & 26.30 & 34.81 \\

&Source bridging  & 36.79\ddag & 38.71\ddag & 37.24\ddag & 40.28\ddag & 27.40\ddag & 35.91\\

&Target bridging  & 36.69  & 39.04\ddag  & 37.63\ddag  & 40.41\ddag & \textbf{27.98}\ddag & 36.27 \\

&Direct bridging  & \textbf{36.97}\ddag & \textbf{39.77}\ddag & \textbf{38.02}\ddag & \textbf{40.83}\ddag & 27.85\ddag & \textbf{36.62}\\
\hline
\hline
\multirow{5}{*}{TER} & cdec (SMT)               & 58.29& 59.65  & 59.28  & 58.12  & 61.54  & 59.64  \\
\cline{2-8}
&RNNSearch*         & 59.56 & 57.79  & 59.25& 57.88  & 64.22  & 59.78\\

&Source bridging  & 58.13 & \textbf{56.25} & 57.33 & 56.32 & 62.13 & \textbf{58.01}\\

&Target bridging  & 58.01  & 56.27  & 57.76 & 56.33 & \textbf{62.12} & 58.12 \\

&Direct bridging  & \textbf{57.20} & 56.68 & \textbf{57.29} & \textbf{55.62} & 62.49 & 58.02\\
\hline
\end{tabular}
\caption{\label{font-table1} BLEU and TER scores on the NIST Chinese-English translation tasks. The BLEU scores are case-insensitive. Avg means the average scores on all test sets. ``\ddag'': statistically better than RNNSearch* (p \(<\) 0.01).  Higher BLEU (or lower TER) scores indicate better translation quality.}
\end{table*}

\subsection{Experimental Settings}

We resorted to Chinese-English bilingual corpora that contain 1.25M sentence pairs extracted from LDC corpora, with 27.9M Chinese words and 34.5M English words respectively.\footnote{ The corpora include LDC2002E18, LDC2003E07, LDC2003E14, Hansards portion of LDC2004T07, LDC2004T08 and LDC2005T06.} We chose the NIST06 dataset as our development set, and the NIST02, NIST03, NIST04, NIST08 datasets as our test sets.

We used the case-insensitive 4-gram NIST BLEU score as our evaluation metric \cite{papineni2002bleu} and the script  `mteval-v11b.pl' to compute BLEU scores. We also report TER scores on our dataset \cite{snover2006ter}.

For the efficient training of the neural networks, we limited the source (Chinese) and target (English) vocabularies to the most frequent 30k words, covering approximately 97.7\% and 99.3\% of the two corpora respectively. All the out-of-vocabulary words were mapped to the special token \textit{UNK}. 
The dimension of word embedding was 620 and the size of the hidden layer was 1000. All other settings were the same as in \citeauthor{bahdanau2015neural} \shortcite{bahdanau2015neural}.
The maximum length of sentences that we used to train the NMT model in our experiments was set to 50, for both the Chinese and English sides. 
Additionally, during decoding, we used the beam-search algorithm and set the beam size to 10. The model parameters were selected according to the maximum BLEU points on the development set. 

We compared our proposed models against the following two systems:
\begin{itemize}
\item cdec \cite{dyer2010cdec}: this is an open source hierarchical phrase-based SMT system \cite{chiang2007hierarchical} with default configuration and a 4-gram language model trained on the target side of the training data.

\item {\bf RNNSearch*}: this is an attention-based NMT system, taken from the dl4mt tutorial with slight changes. It improves the attention model by feeding the lastly generated word. For the activation function \(f\) of an RNN, we use the gated recurrent unit (GRU) \cite{chung2014empirical}. Dropout was applied only on the output layer and the dropout \cite{hinton2012improving} rate was set to 0.5. We used the stochastic gradient descent algorithm with mini-batch and Adadelta \cite{zeiler2012adadelta} to train the NMT models. The mini-batch was set to 80 sentences and decay rates \(\rho\) and \(\varepsilon\) of Adadelta were set to 0.95 and \(10^{-6}\). 
\end{itemize}

For our NMT system with the direct bridging model, we use a simple pre-training strategy to train our model. We first train a regular attention-based NMT model, then use this trained model to initialize the parameters of the NMT system equipped with the direct bridging model and randomly initialize the additional parameters of the direct bridging model in this NMT system. The reason of using pre-training strategy is that the embedding loss requires well-trained word alignment as a starting point.

\subsection{Experimental Results}

Table~\ref{font-table1} displays the translation performance measured in terms of BLEU and TER scores. Clearly, every one of the three NMT models we proposed, with some bridging method, improve the translation accuracy over all test sets in comparison to the SMT (cdec) and NMT (RNNSearch*) baseline systems.

\subsubsection*{Parameters}

The three proposed models introduce new parameters in different ways. The source-side bridging model augments source hidden states from a dimension of 2,000 to 2,620, requiring 3.7M additional parameters to accommodate the  hidden states that are appended. 
The target-side bridging model introduces 1.8M additional parameters for catering $x_{t^*}$ in calculating target side state, as in Equation 1.  
Based on the source-side bridging model, the direct bridging model requires extra 0.4M parameters (i.e. the transformation matrix $W$ in Equation 3 is $620 * 620$), resulting in 4.1M additional parameters over the baseline. 
Given that the baseline model has 74.8M parameters, the size of extra parameters in our proposed models are comparably small.

\subsubsection*{Comparison with the baseline systems} 
The results in Table~\ref{font-table1} indicate that all NMT systems outperform the SMT system taking into account the evaluation metrics considered, BLEU and TER. This is consistent with other studies on Chinese to English machine translation \cite{mi2016supervised,tu2016modeling,li2017modeling}. 
Additionally, all the three NMT models with bridging mechanisms we proposed outperform the baseline NMT model RNNSearch*. 

With respect to BLEU scores, we observe a consistent trend that the target-side bridging model works better than the source-side bridging model, while the direct bridging model achieves the best accuracy over all test sets, with the only exception of NIST MT 08. On all test sets, the direct bridging model outperforms the baseline RNNSearch* by 1.81 BLEU points and outperforms the other two bridging-improved NMT models by 0.4$\sim$0.6 BLEU points. 

Though all models are not tuned on TER score, our three models perform favorably well with similar average improvement, of about 1.70 TER points, below the baseline model. 

\begin{table}[!t]
\centering
\small
\begin{tabular}{cc}
\hline
\bf System & \bf Percentage (\%) \\
\hline
RNNSearch*         & 49.82  \\
\hline
Direct bridging        & 81.30   \\
\hline
\end{tabular}
\caption{\label{font-table2} Percentage of target side \textit{eos} translated from source side \textit{eos} on the development set.}
\end{table}

\begin{table*}[!t]
\centering
\small
\begin{tabular}{c|c|cccccccc}
\hline
\multirow{2}{*}{\bf System} & \multirow{2}{*}{\bf Target POS Tag} & \multicolumn{5}{c}{\bf Source POS Tag}\\
\cline{3-7}
& & \bf V  & \bf N  & \bf CD & \bf JJ  & \bf AD \\
\hline
\multirow{4}{*}{RNNSearch*} 
& V	& 64.95        &  -   & -      & -       & 12.09  \\
& N	& 7.31    &  39.24    & -      & -       & -      \\ 
& CD	&  -   & 33.37        & 53.40      & -   & -      \\
& JJ    & -    & 26.79        & -  & 14.67       & -       \\
\hline
\hline
\multirow{4}{*}{Direct bridging} 
&V	    & 66.29        &  -  & -      & -       & 10.94  \\
&N	    &  7.19   & 39.71   & -      & -       & -      \\
&CD	    &  -   & 32.25       & 56.29      & -   & -      \\
&JJ	    & -    & 26.12       & -  & 15.22       & -      \\
\hline
\end{tabular}
\caption{\label{font-table3} Confusion matrix for translation by POS, in percentage. To cope with fine-grained differences among verbs (e.g., VV, VC and VE in Chinese, and VB, VBD, VBP, etc. in English), we merge all verbs into V. Similarly, we merged all nouns into N. CD stands for Cardinal numbers, JJ for adjectives or modifiers, AD for adverbs. These POS tags exist in both Chinese and English. For the sake of simplicity, for each target POS tag, we present only the two source POS tags that are more frequently aligned with it.}
\end{table*}

\section{Analysis}
As the direct bridging system proposed achieves the best performance, we further look at it and at the RNNSearch* baseline system to gain further insight on how bridging may help in translation. Our approach presents superior results along all the dimensions assessed.

\subsection{Analysis of Word Alignment}
Since our improved model strengthens the focus of attention between pairs of translation equivalents by explicitly bridging source and target word embeddings, we expect to observe improved word alignment quality. The quality of the word alignment is examined from the following three aspects.

\subsubsection*{Better \textit{eos} translation}  As a special symbol marking the end of sentence, target side \textit{eos} has a critical impact on controlling the length of the generated translation. 
A target \textit{eos} is a correct translation when is aligned with the source \textit{eos}. 
Table \ref{font-table2} displays the percentage of target side \textit{eos} that are translations of the source side \textit{eos}. It indicates that our model improved with bridging substantially achieves better translation of source \textit{eos}. 

\subsubsection*{Better word translation} To have a further insight into the quality of word translation, we group generated words by their part-of-speech (POS) tags and examine the POS of their aligned source words.~\footnote{We used Stanford POS tagger~\cite{toutanova2003tagger} to get POS tags for the words in source sentences and their translations.} 

Table~\ref{font-table3} is a confusion matrix for translations by POS. For example, under RNNSearch*, 64.95\% of target verbs originate from verbs in the source side. This is enhanced to 66.29\% in our direct bridging model. 
From the data in that table, one observes that in general more target words align to source words with the same POS tags in our improved system than in the baseline system. 

\begin{table}[!t]
\centering
\small
\begin{tabular}{ccc}
\hline
\bf System & \bf SAER & \bf AER \\
\hline
RNNSearch*        & 62.68  & 47.61 \\
\hline


Direct bridging        & 59.72   & 44.71  \\
\hline
\end{tabular}
\caption{\label{font-table4} Alignment error rate (AER) and soft AER. quality. A lower score indicates better alignment.}
\end{table}

\subsubsection*{Better word alignment} 
Next we report on the quality of word alignment taking into account a manually aligned dataset, from \citeauthor{liu2015contrastive}~\shortcite{liu2015contrastive}, which contains 900 manually aligned Chinese-English sentence pairs. We forced the decoder to output reference translations in order to get automatic alignments between input sentences and their reference translations yielded by the translation systems. To evaluate alignment performance, we measured the alignment error rate (AER) \cite{och2003systematic} and the soft AER (SAER) \cite{tu2016modeling}, which are registered in Table~\ref{font-table4}.

The data in this Table 4 indicate that, as expected, bridging improves the alignment quality as a consequence of its favoring of a stronger relationship between the source and target word embeddings of translational equivalents.

\begin{figure}[!t]
\centering
\includegraphics*[height=1.2in,width=2.2in]{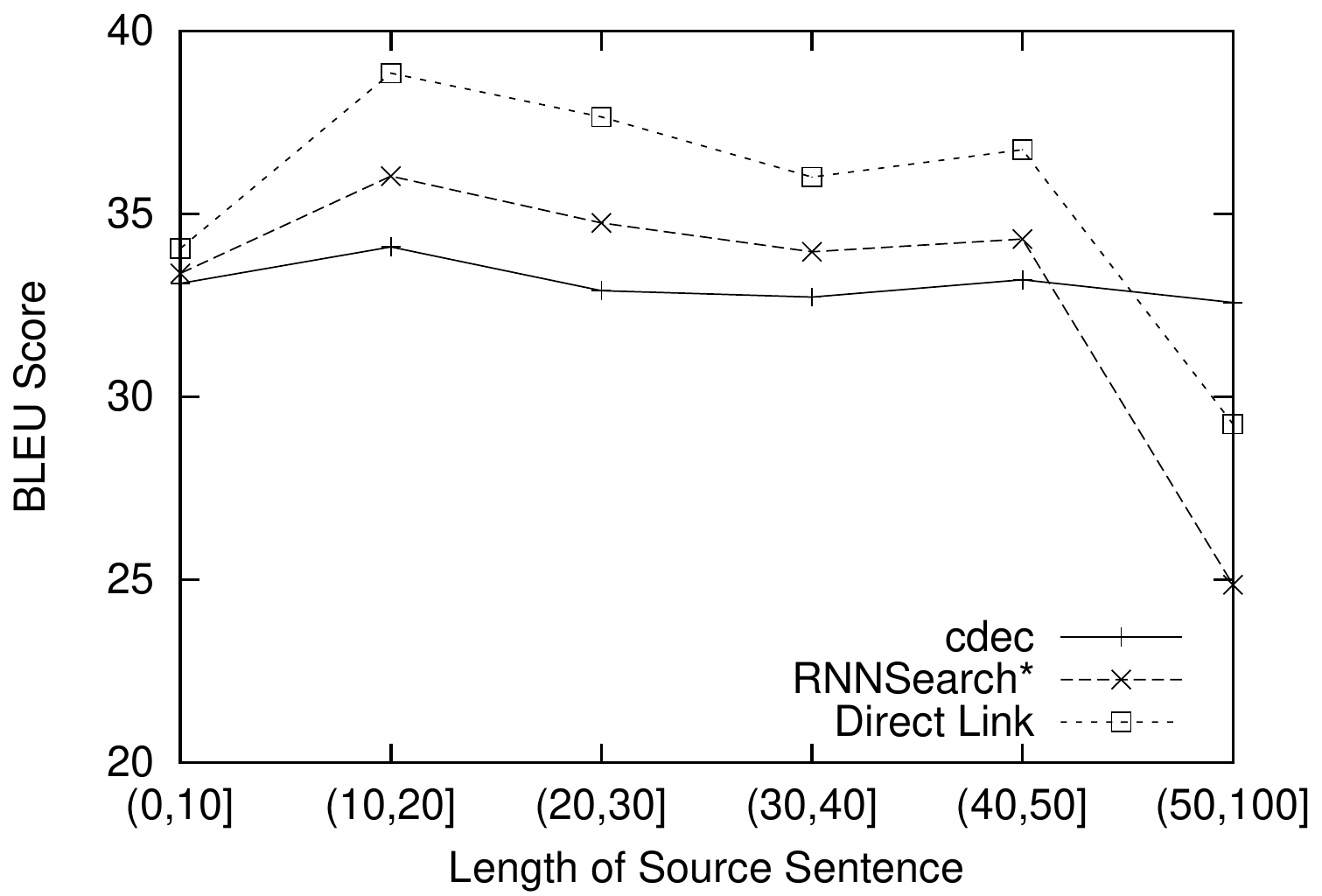}
\caption{BLEU scores for the translation of sentences with different lengths.}
\label{fig:5}
\end{figure}

\subsection{Analysis of Long Sentence Translation}

Following~\citeauthor{bahdanau2015neural}~\shortcite{bahdanau2015neural}, we partition sentences by their length and compute the respective BLEU scores, which are presented in Figure~\ref{fig:5}. These results indicate that our improved system outperforms RNNSearch* for all the sentence lengths. 
They also reveal that the performance drops substantially when the length of the input sentence increases.  This  trend is consistent with the findings in \cite{cho2014learning,tu2016modeling,li2017modeling}. 

One also observes that the NMT systems perform very badly on sentences of length over 50, when compared to the performance of the baseline SMT system (cdec). We think that the degradation of NMT systems performance over long sentences is due to the following reasons: (1) during training, the maximum source sentence length limit is set to 50, thus making the learned models not ready to cope well with sentences over this maximum length limit; (2) for long input sentences, NMT systems tend to stop early in the generation of the translation.

\subsection{Analysis of Over and Under Translation}

To assess the expectation that improved translation of \textit{eos} improves the appropriate termination of the translations generated by the decoder, we analyze the performance of our best model with respect to over translation and under translation, which are both notoriously a hard problem for NMT.

To estimate the over translation generated by an NMT system, we follow \citeauthor{li2017modeling}~\shortcite{li2017modeling} and report the ratio of over translation (ROT)\footnote{please refer to \cite{li2017modeling} for more details of ROT.}, which is computed as the total number of times of over translation of words in a word set (e.g., all nouns in the source part of the test set) divided by the number of words in the word set.

Table~\ref{font-table5} displays ROTs of words grouped by some prominent POS tags. These data indicate that both systems have higher over translation with proper nouns (NR) and other nouns (NN) than with other POS tags, which is consistent with the results in \cite{li2017modeling}. The likely reason is that these two POS tags usually have more unknown words, which are words that tend to be over translated. 
Importantly, these data also show that our direct bridging model alleviates the over translation issue by 15\%, as ROT drops from 5.28\% to 4.49\%.
\ignore{
\begin{equation}
ROT = \frac{\sum_{w_i} t(w_i)}{|w|}
\end{equation}
\\where \(|w|\) is the number of source words under consideration, \(t(w_i)\) is the number of times of over translation for word \(w_i\), which is given by:

\begin{equation}
t(w) = |e| - |uniq(e)|
\end{equation}
\\where \(|e|\) is the number of words in the translation \(e\) of \(w\), while \(|uniq(e)|\) is the number of unique words in \(e\).
}

\begin{table}[!t]
\centering
\small
\begin{tabular}{c|cc}
\hline
\bf System & POS & \bf ROT($\%$) \\
\hline
\multirow{5}{*}{RNNSearch*} 
& NN        & 8.63  \\
& NR        & 12.92 \\
& DT        & 4.01  \\
& CD        & 7.05  \\
\cline{2-3}
& ALL       & 5.28  \\
\hline
\hline

\multirow{5}{*}{Direct bridging} 
& NN        & 7.56  \\
& NR        & 10.88 \\
& DT        & 2.37  \\
& CD        & 4.79  \\
\cline{2-3}
& ALL       & 4.49  \\
\hline
\end{tabular}
\caption{\label{font-table5} Ratios of over translation (ROT) on test sets. NN stands for nouns excluding proper nouns and temporal nouns, NR for proper nouns, DT for determiners, and CD for cardinal numbers.}
\end{table}

\begin{table}[!t]
\centering
\small
\begin{tabular}{ccc}
\hline
\bf System & \bf 1-gram BLEU \\
\hline
cdec (SMT) & 77.09 \\
\hline
RNNSearch*        & 72.70  \\
Direct bridging        & 74.22  \\
\hline
\end{tabular}
\caption{\label{font-table6} 1-gram BLEU scores averaged on test sets, supporting the assessment of under translation. A larger score indicates less under translation.}
\end{table}

While it is hard to obtain an accurate estimation of under translation, we simply report 1-gram BLEU score that measures how many words in the translation outcome appear in the reference translation, roughly indicating the proportion of source words that are translated. 
Table~\ref{font-table6} presents the average 1-gram BLEU scores on our test datasets. These data indicate that our improved system has a higher score than RNNSearch*, suggesting that it is less prone to under translation. 

It is also worth noting that the SMT baseline (cdec) presents the highest 1-gram BLEU score, as expected, given that under translation is known to be less of an issue for SMT.

\subsection{Analysis of Learned Word Embeddings}
In the direct bridging model, we introduced a transformation matrix to convert a source-side word embedding into its counterpart on the target side. We seek now to assess the contribution of this transformation. 
Given a source word $x_i$, we obtain its closest target word $y^*$ via:

\begin{equation}
y^* = \operatorname{arg\,min}_{y}(\|wx_i - y\|)
\end{equation}

Table~\ref{font-table7} lists the 10 more frequent source words and their corresponding closest target words. For the sake of comparison, it also displays their most likely translations from the lexical translation table in SMT. These results suggest that the closest target words obtained via the transformation matrix of our direct bridging method are very consistent with those obtained from the SMT lexical table, containing only admissible translation pairs. These data thus suggest that our improved model has a good capability of capturing the translation equivalence between source and target word embeddings.

\begin{table}[!t]
\centering
\small
\begin{tabular}{c|cc}
\hline
\bf Src & \bf Transformation & \bf Lexical Table  \\
\hline
是         & is         & is      \\
\hline
和  & and   & and  \\
\hline
及  & and   & and  \\
\hline
将  & will  &  will \\
\hline
会  & will  &  will \\
\hline
国         & countries  & countries   \\
\hline
发展       & development & development   \\
\hline   
经济 & economic & economic \\
\hline
问题 & question & issue \\
\hline
人民 & people & people \\
\hline
\end{tabular}
\caption{\label{font-table7} Top 10 more frequent source words and their closest translations obtained,  respectively, by embedding transformation in NMT and from the lexical translation table in SMT.}
\end{table}

\section{Related Work}
Since the pioneer work of \citeauthor{bahdanau2015neural} \shortcite{bahdanau2015neural} to jointly learning alignment and translation in NMT, many effective approaches have been proposed to further improve the alignment quality. 

The attention model plays a crucial role in the alignment quality and thus its enhancement has continuously attracted further efforts. To obtain better attention focuses, \citeauthor{luong2015effective}~\shortcite{luong2015effective} propose global and local attention models; and \citeauthor{cohn2016biase}~\shortcite{cohn2016biase} extend the attentional model to include structural biases from word based alignment models, including positional bias, Markov conditioning, fertility and agreement over translation directions. 

In contrast, we did not delve into the attention model or sought to redesign it in our new bridging proposal. And yet we achieve enhanced alignment quality by inducing the NMT model to capture more favorable pairs of words that are translation equivalents of each other under the effect of the bridging mechanism.

Recently there have been also studies towards leveraging word alignments from SMT models. \citeauthor{mi2016supervised} \shortcite{mi2016supervised} and \citeauthor{liu2016neural}~\shortcite{liu2016neural} use pre-obtained word alignments to guide the NMT attention model in the learning of favorable word pairs. \citeauthor{arthur2016lexicons}~\shortcite{arthur2016lexicons} leverage a pre-obtained word dictionary to constrain the prediction of target words. 
Despite these approaches having a somewhat similar motivation of using pairs of translation equivalents to benefit the NMT translation, in our new bridging approach we do not use extra resources in the NMT model, but let the model itself learn the similarity of word pairs from the training data.~\footnote{Though the pre-obtained word alignments or word dictionaries can be learned from MT training data in an unsupervised fashion, these are still extra knowledge with respect to to the NMT models.}

Besides, there exist also studies on the learning of cross-lingual embeddings for machine translation. \citeauthor{mikolov2013exploiting}~\shortcite{mikolov2013exploiting} propose to first learn distributed representation of words from large monolingual data, and then learn a linear mapping between vector spaces of languages. \citeauthor{gehring2017convolutional}~\shortcite{gehring2017convolutional} introduce source word embeddings to predict target words. 
These approaches are somewhat similar to our source-side bridging model. However, inspired by the insight of shortening the distance between source and target embeddings in the seq2seq processing chain, in the present paper we propose more strategies to bridge source and target word embeddings and with better results.

\section{Conclusion}


We have presented three models to bridge source and target word embeddings for NMT.
The three models seek to shorten the distance between source and target word embeddings along the extensive information procedure in the encoder-decoder neural network. 

Experiments on Chinese to English translation shows that the proposed models can significantly improve the translation quality. Further in-depth analysis demonstrate that our models are able (1) to learn better word alignments than the baseline NMT, (2) to alleviate the notorious problems of over and under translation in NMT, and (3) to learn direct mappings between source and target words. 

In future work, we will explore further strategies to bridge the source and target side for sequence-to-sequence and tree-based NMT. Additionally, we also intend to apply these methods to other sequence-to-sequence tasks, including natural language conversation.

\section*{Acknowledgment} 

The present research was partly supported 
by the National Natural Science
Foundation of China (Grant No. 61622209), the CNPTDeepMT grant of the Portugal-China Bilateral Exchange Program (ANI/3279/2016) 
and the Infrastructure for the Science and Technology of the Portuguese Language (PORTULAN / CLARIN).

\bibliography{acl2018}
\bibliographystyle{acl_natbib}

\end{CJK}
\end{document}